Stefan Strecker, Jürgen Jung (Eds.)

# Informing Possible Future Worlds

Essays in Honour of Ulrich Frank

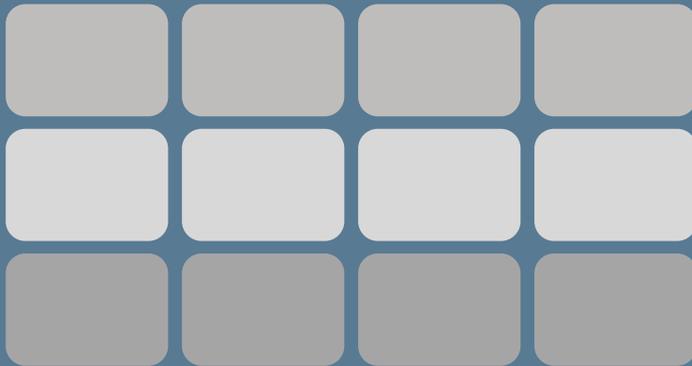

Aspects

Perspectives

λογος

Chapter 15

# Enterprise Use Cases Combining Knowledge Graphs and Natural Language Processing

Phillip Schneider, Tim Schopf, Juraj Vladika and Florian Matthes

Knowledge management is a critical challenge for enterprises in today's digital world, as the volume and complexity of data being generated and collected continue to grow incessantly. Knowledge graphs (KG) emerged as a promising solution to this problem by providing a flexible, scalable, and semantically rich way to organize and make sense of data. This paper builds upon a recent survey of the research literature on combining KGs and Natural Language Processing (NLP). Based on selected application scenarios from enterprise context, we discuss synergies that result from such a combination. We cover various approaches from the three core areas of KG construction, reasoning as well as KG-based NLP tasks. In addition to explaining innovative enterprise use cases, we assess their maturity in terms of practical applicability and conclude with an outlook on emergent application areas for the future.

## 15.1 Introduction

As modern organizations continuously adapt to the evolving requirements of the digital age, the importance of successfully managing enterprise data has never been greater. In this article, we define the term 'enterprise' as a large-scale business, which operates on a national or international level and typically involves significant risks and resources. The competitive advantage of data-driven decisions affects all industry sectors and nearly every part of the value chain (Schopf et al. 2022b). For example, market trends are analyzed for business development, production is optimized through process metrics, and customer reviews are monitored for predictive maintenance. However, raw data alone is insufficient for decision-making. In order to become actionable information, data has to be endowed with meaning and purpose (Rowley 2007). This can be achieved by data enrichment through a relevant context. A compelling approach to achieve this is by modeling knowledge in the form of graph connections between data items (Martin et al. 2021).

In view of the above, knowledge graphs (KGs) have emerged as a powerful representation for integrating knowledge from multiple information sources. They model semantic relationships among key entities of interest, such as customers, markets, or services. In this sense, a KG serves as an abstraction layer to combine both business data and explicit business knowledge. Even though Google coined the term knowledge graph back in 2012 when they announced their new web search strategy (Singhal 2012), it is not an entirely novel concept. The notion of knowledge graphs stems from scientific advancements in areas like semantic web, database systems, and machine learning. Fueled by the demand for intelligent data management and strong enterprise use cases, the integration of ideas from the aforementioned fields has attracted a surge of interest from both academia and industry in the last years (Hitzler 2021).

In particular, the adoption of KGs has been noticeable in the field of natural language processing (NLP). The underlying paradigm is that graphs can serve as an intermediate formalism to bridge structured and unstructured, textual knowledge. This entails potential benefits for all kinds of NLP tasks. Because a major part of business knowledge resides in unstructured text sources, leveraging KGs and NLP for data management appears to be an auspicious endeavor. In our comprehensive survey on the research landscape of KGs in NLP, we provide an overview of the rapidly expanding body of literature (Schneider et al. 2022a). Parallel to this trend, the variety of explored use cases and application domains grew as well. We identified 20 individual domains, many of which belong to enterprise-related subdomains. Yet, despite numerous scientific contributions, there are very few publications discussing the promised usefulness in practice. Therefore, the following research question arises:

*What is the added value of combining KGs and NLP for enterprise use cases?* Taking a step towards answering this research question, we build upon our survey results and investigate the synergies between KGs and NLP with selected enterprise use cases, ranging from knowledge acquisition to knowledge application. Concrete examples are discussed that showcase the added value that this form of graph-based knowledge processing brings. Besides explaining innovative methods for the three core areas of KG construction, KG reasoning, and KG-based NLP applications, we also discuss their maturity in terms of practical applicability. We conclude with an outlook on particularly promising KG and NLP application scenarios for future research.

## 15.2 Background and Terminology

### 15.2.1 Knowledge Graph Concept

In recent years, KGs have emerged as an approach for semantically representing knowledge about real-world entities in a machine-readable format. In contrast to semantic text representations, which may be used primarily for similarity comparisons in a variety of different NLP downstream tasks (Braun et al. 2021; Schopf et al. 2021a; Schopf et al. 2021b; Schopf et al. 2022d; Schopf et al. 2022c; Schopf et al. 2022a; Schneider et al. 2022b), KGs can additionally capture all kinds of semantic relationships between different entities. Despite the rising popularity of KGs, there is still no common understanding of what exactly a KG is. Although prior work has already attempted to define KGs (Pujara et al. 2013; Ehrlinger and Wöß 2016; Paulheim 2017; Färber et al. 2018), the term is not yet used uniformly by researchers. Most works implicitly adopt a broad definition of KGs, where they are understood as *'a graph of data intended to accumulate and convey knowledge of the real world, whose nodes represent entities of interest and whose edges represent relations between these entities'* (Hogan et al. 2022). KGs originated from research on semantic networks, domain-specific ontologies, as well as linked data, and are thus not an entirely new concept (Hitzler 2021). Since the concepts of ontologies, semantic networks, and KGs are strongly related, these terms are often mistakenly used as synonyms, leading to additional confusion (Ehrlinger and Wöß 2016). Although an ontology can also hold instances (i. e., the population of the ontology), an ontology mainly consists of classes and properties as a formal, explicit specification of a shared conceptualization that is characterized by high semantic expressiveness required for increased complexity (Feilmayr and Wöß 2016). Semantic networks can be seen as predecessors of KGs (Hogan et al. 2022) and represent semantic relations between conceptual units as a net-like graph ('Semantic networks' 1992). One of the most well-known semantic networks in NLP is WordNet, which consists of English nouns, verbs, adjectives, and adverbs that are organized into sets of synonyms, each representing a lexicalized concept (Miller 1995).

Relational Database Management Systems (RDBMS) are collections of data that are stored in manually modeled tables that are connected to each other (Medhi and Baruah 2017). In contrast to KGs, RDBMS do not store relationships between individual instances (nodes) but use predefined tabular schemas. Therefore, RDBMS are more powerful when processing large amounts of data that do not change in structure. KGs, on the other hand, are more powerful when data is highly connected, the data model changes frequently, or data retrieval and analysis are more important than simply storing the data. Since companies often have multiple databases that are semantically related and also require analysis or further processing of the data for downstream NLP tasks, the current tendency is to create enterprise knowledge graphs.

Enterprise knowledge graphs consist of connected concepts, properties, instances and relationships representing and referencing foundational and domain knowledge within or across different enterprises (Fraunhofer IAIS 2021). They are realizations of linked enterprise data, creating valuable business insights by bringing together semantic technologies and enterprise data infrastructures (Fraunhofer IAIS 2021). Enterprise knowledge graphs are typically internal to a company and applied for commercial use-cases (Noy et al. 2019). Prominent industries using enterprise knowledge graphs include web search, e-commerce, social networks, and finance, among others, where applications include search, recommendations, information extraction, personal agents, advertising, business analytics, risk assessment, automation, and more (Hogan et al. 2022).

### 15.2.2 Knowledge Graphs in Natural Language Processing

NLP is a field at the intersection of linguistics, computer science, and artificial intelligence. Its goal is to build computer systems that can understand and interact with human language and solve various tasks involving language data. The way humans communicate and exchange knowledge is often in an unstructured format – large corpora of written text documents, presentation slides, e-mails, or chat messages. There is a trend of trying to utilize the structured representation of knowledge found in KGs in combination with the knowledge in unstructured text. KGs are a useful tool for NLP because they provide a structured, machine-readable representation of real-world entities and the relationships between them. This allows NLP systems to understand the meaning of words and phrases in context, and to use this understanding to answer questions, generate text summaries, and perform other knowledge-intensive tasks. Additionally, KGs can be used to improve the accuracy and performance of NLP systems by providing them with a wealth of background information about the world that they can use to disambiguate words and phrases and make more informed decisions.

## 15.3 Analysis of Enterprise Use Cases

Based on the tasks identified in the literature on KGs in NLP, we developed the empirical taxonomy shown in Figure 15.1. If one or multiple of these tasks are applied to a specific domain with the goal to create business value, we consider it as an enterprise use case. In this regard, one can distinguish between two top-level categories, namely knowledge acquisition and knowledge application. Knowledge acquisition contains NLP tasks to construct KGs from unstructured text (knowledge graph construction) or to conduct reasoning over already constructed KGs (knowledge graph reasoning). KG construction tasks are further split into the subcategories of knowledge extraction and knowledge integration. KG reasoning is used to maintain and update facts inside KGs. Lastly, knowledge application, being the second top-level category, encompasses common NLP tasks, which can be enhanced through structured knowledge from KGs.

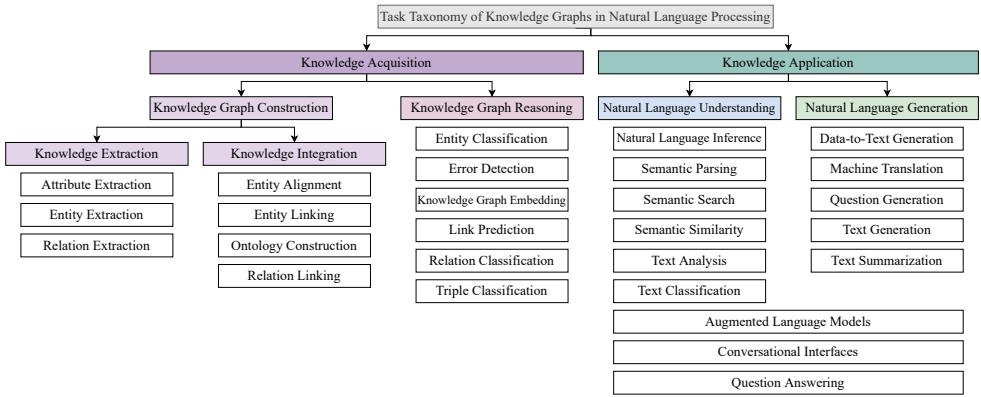

Figure 15.1: Taxonomy of tasks in the literature on KGs in NLP (Schneider et al. 2022a).

In the following subchapters, the aforementioned tasks combining KGs and NLP are explained in more detail. To demonstrate the potential value of such a combination, we introduce selected studies that exemplify enterprise use cases across multiple industry sectors. Table 15.1 provides an overview of additional studies for the described use cases.

### 15.3.1 Knowledge Graph Construction

Knowledge graph construction in NLP considers tasks which involve the explicit modeling of knowledge in KGs from unstructured text. As shown in Figure 15.1, KG construction can be further divided into the two subcategories of knowledge extraction and knowledge integration. Knowledge extraction tasks are used to populate KGs with entities, relations, or attributes, whereas knowledge integration tasks are used to update KGs (Schneider et al. 2022a).

Usually, *entity extraction* is the first step in the KGs construction process and is used to derive real-world entities from unstructured text (Al-Moslmi et al. 2020). After the entities are extracted, their relationships can be extracted with the task of *relation extraction* (Zhang et al. 2019). Entity extraction in combination with relation extraction is often used to construct new KGs, e. g., for news events (Rospocher et al. 2016), scholarly research (Luan et al. 2018; Shen et al. 2018), patent mining (Wu et al. 2021), engineering technologies (Sarica et al. 2020), or the petroleum industry (Zhou et al. 2020). *Entity linking* tasks link entities recognized in a text to already existing entities in KGs (Moon et al. 2018; Wu et al. 2020). Oftentimes, synonymous or similar entities exist in different KGs. To reduce redundancy and repetition in future tasks, *entity alignment* can be applied to align these entities (Gangemi et al. 2016; Chen et al. 2018). Coming up with the rules and schemes of KGs, i. e., their structure and format of knowledge presented in it, is done with the task of *ontology construction* (Haussmann et al. 2019; Schneider et al. 2022a). In the following, we present two selected KGs constructed from texts which support knowledge-intensive processes in enterprises of different industries.

| Enterprise Use Cases | Identified Domains | Selected Studies |
| --- | --- | --- |
| Analysis & monitoring | Operations, tech. scouting, reporting | Gua and Liu 2019; Qin and Chow 2019; Braaksma et al. 2021; Hu et al. 2021 |
| Digital assistant | Finance, human resources | Singh et al. 2018; Avila et al. 2020; McCrae et al. 2021 |
| Document generation | Scholarly publishing | Fan et al. 2019; Koncel-Kedziorski et al. 2019; Wang et al. 2021 |
| Knowledge repository | Engineering, scholarly publishing | Li et al. 2020; Sarica et al. 2020; Wang et al. 2020 |
| Search engine | Energy, engineering, law, software | Birnie et al. 2019; Sarica et al. 2019; Zhou et al. 2020; Gao et al. 2021 |
| Trade recommendation | Finance, law | Ding et al. 2016; Du et al. 2022 |

Table 15.1: Selected studies of enterprise use cases in the literature on KGs in NLP.

*Microsoft Academic Graph.* The average growth rate ranged from 5 to 6.5 per cent for scientific articles and from 2 to 3 per cent for journals between 2015 and 2020, causing the number of publications to grow exponentially over the past centuries (International Association of Scientific, Technical and Medical Publishers 2021). Due to this unprecedented growth of scientific knowledge, efficiently identifying relevant research has become an ever-increasing challenge (Shen et al. 2018). Enterprises often seek to increase their capabilities and innovation by incorporating new scientific discoveries into their business processes. In order to assist researchers and businesses to navigate the entirety of scientific information, the Microsoft Academic Graph (MAG) organizes scientific knowledge in a hierarchical manner (Shen et al. 2018). MAG is a KG in which the nodes represent scientific entities and the edges represent the relationships between them (Wang et al. 2020). In the MAG, publications are the main entity type to which all other entities (e. g., authors, venues, field of study concepts, etc.) are connected to in one way or another (Wang et al. 2020).

*TechNet.* Various NLP techniques were used to extract terms and their relationships from the complete 1976–2017 U.S. patent database and to create TechNet (Sarica et al. 2020). It is a semantic network that covers elemental concepts in all domains of technology and their semantic associations. TechNet may serve as an infrastructure to support a wide range of applications, e. g., technical text summaries, search query predictions, relational knowledge discovery, and design ideation support, in the context of engineering and technology, and complement or enrich existing semantic databases. Therefore, TechNet can support corporations with an up-to-date resource for technical knowledge and help them stay competitive in a rapidly changing technological landscape.

### 15.3.2 Knowledge Graph Reasoning

Subsequent to the construction of a KG, it contains world knowledge in a structured format. This knowledge can be general encyclopedic knowledge or domain-specific knowledge that is useful for a certain target group. The tasks in the group of KG reasoning aim to infer new knowledge and conclusions by using the already present knowledge in the KG.

Written text often contains named entities and concepts that have to be categorized in order for NLP techniques to better understand them. The task of classifying entities found in the text based on the knowledge from a KG is called *entity classification*, while the task of retrieving a relation from a KG between two entities found in text is called *relation classification*. If two entities and their relation is considered jointly in a triple format, this task is called *triple classification*. The task of *link prediction* tries to infer missing links between entities in existing KGs and is often performed via ranking entities as possible answers to queries. It is also possible that there are erroneous entities or relations in a KG – the task of *error detection* addresses this problem and aims to refine the knowledge of a KG. *Knowledge graph embedding* techniques are used to create dense vector representations of knowledge from the graph.

Misinformation detection is a NLP task where KG reasoning and external knowledge are important for enterprises. This task has gained wide attention because the spread of fake news and biased news can have significant socio-economic consequences, increase mistrust within society, and lead to stock market perturbations or other economic disruptions (Scheufele and Krause 2019; Palić et al. 2019; Clarke et al. 2021). Some approaches for fake news detection rely solely on linguistic cues such as how sensationalist or persuasive the text of a news article is, or on social cues like where the news first originated and how it was propagated. Nevertheless, for optimal performance, NLP models for this task need to incorporate background knowledge for a deeper understanding of what is being described in the text. For example, Hu et al. (2021) match entities such as public persons, places, or events mentioned in news articles to their respective entities in a knowledge graph and then train graph neural networks (GNNs) to learn about the context of each mentioned entity. This new knowledge is then used to enhance the performance of fake news detection.

Most KG reasoning applications rely on representing knowledge graph concepts as embeddings which are then in a suitable format to be used as input to machine learning models like deep neural networks. This concept is also used in NLP to represent words as vector embeddings. Sometimes text and graph embeddings are learned jointly, as to make their representation closer in the embedding space. In Ding et al. (2016), the authors propose a model that jointly learns event embeddings by using both the textual and world knowledge from a KG. The model utilizes world knowledge from the YAGO knowledge graph, which was constructed using knowledge from Wikipedia (Suchanek et al. 2007). Examples of triples in that KG include *(Bill Gates, created, Microsoft)* and *(Apple, isA, electronics company)*. Afterward, the authors use these event embeddings for the task of stock market prediction and show that such joint embeddings achieved better performance on the task of stock market volatility prediction.

### 15.3.3 Knowledge-Based Natural Language Understanding and Generation

After the construction and reasoning-enabled refinement of a KG, it can be used as an information source for enhancing tasks in the areas of natural language understanding and natural language generation. While the former is concerned with parsing syntax, semantics, or pragmatics and thus understanding the meaning of natural language, the latter focuses on producing human-like text based on linguistic or non-linguistic input.

Automated approaches for extracting the content and meaning of textual data are of great importance for enterprises. Having pertinent information at hand is the foundation of data-driven decision-making. In this regard, the main challenge of managing knowledge lies not in storing data in digital repositories, but in retrieving the relevant information pieces amongst an overwhelming mass of other data items. One of the key tasks of using KGs in natural language understanding is *semantic search*, which refers to 'search with meaning'. Semantic search goes beyond literal keyword matches and aims at understanding the search intent and context (Bast et al. 2016). This increases the information retrieval performance despite ambiguous or poorly phrased queries. Semantic search is often a part of business information systems like search engines, recommender systems, or analytical tools. Aside from understanding language, KGs can be used for the task of *text generation*. This concerns automatically generating business reports or other textual outputs that provide a concise representation of the enterprise data contained in a KG.

Bringing together language understanding and generation capabilities, *question answering (QA)* is another common NLP task. Its goal is to find an answer given a user query formulated in natural language. QA systems can be distinguished into three categories: (1) textual QA for extracting answers from a document corpus, (2) question answering over knowledge bases (KBQA) for getting answers from structured knowledge bases such as KGs (Fu et al. 2020), and (3) hybrid approaches that use a combination of text corpora as well as knowledge bases (Yu et al. 2022). *Conversational interfaces* surpass the restricted question-answer scheme and emulate a realistic dialogue in human language. Users can interact with conversational interfaces through multiple conversation turns, wherefore they are especially suitable to handle more complex information-seeking scenarios (McTear et al. 2016). To name just a few examples, conversational interfaces are commonly used to inform employees about company policies, provide data-based insights for business operations, or improve customer service by quickly resolving issues.

As a semantic search use case related to the previously introduced *TechNet* graph, Sarica et al. (2019) propose a patent retrieval system (*SerKeys*) based on keywords from a constructed engineering knowledge graph. Engineers often search for patents regarding a design topic to learn about the prerequisites of existing technologies. However, discovering patents relevant to a specific design interest, such as additive manufacturing or autonomous logistic robots, is a non-trivial task because keywords are often intuitively chosen and limited by the vocabulary of the searcher. This leads to overlooked keywords, suboptimal queries, and poor search results. To address this concern, the KG-based patent retrieval system lets users select recommended keywords to expand and refine their queries. In a small case study, the authors show that keyword discovery improves the completeness of queries and reduces the search effort of engineers. An additional example of semantic search is *PetroKG*, a KG that integrates heterogeneous data distributed in various sources in the upstream sector of the enterprise PetroChina (Zhou et al. 2020). To search for fossil fuel deposits and extract these resources from the earth, PetroChina has to accumulate a huge amount of production data, online encyclopedia data, and unstructured documents, which are all integrated and semantically interlinked in the PetroKG. The latter is used for a KG-powered semantic search system that detects entities from user input, executes graph queries, and returns concise answers. This knowledge retrieval service supports researchers and engineers employed at PetroChina who previously had to search through long lists of documents and industrial data logs in their daily work.

One example of conversational interfaces is *KNADIA*, an intra-enterprise knowledge-assisted dialogue system that has been deployed in production in Tata Consultancy Services (Singh et al. 2018). Its system architecture supports two use cases: virtual assistance and knowledge synthesis. While the former is about handling questions from employees about human resource policies, the latter aims at storing information about reusable assets, completed projects, or research documents in a KG and making it accessible through a conversational agent. This agent is also able to approach employees with questions to autonomously update the KG. The KNADIA architecture consists of a high-level intent identification layer and a bot manager which matches all user requests to a number of task-specific conversational bots. After its deployment, the creators of KNADIA observed a significant growth of user queries per day, especially subsequent to opening an auto-suggest feature, where the system proactively showed possible questions users could ask based on recent conversations.

Moreover, there are many potential enterprise use cases for generating texts from KGs. Although prior work has demonstrated that language models can produce short texts using structured information from a KG, the existing models are still prone to missing important relations or hallucinating wrong facts (Koncel-Kedziorski et al. 2019; Wang et al. 2021). Strategies for mitigating these problems are a matter of ongoing NLP research. For these reasons, the proposed solutions are almost never deployed and tested in practice.

## 15.4 Discussion

The described studies reveal how enterprises can use KGs as a means of of structuring information and modeling complex relationships. If there is no existing KG to build upon, an enterprise has to go through the mentioned three phases, whereas each phase heavily depends on NLP techniques. The first phase is KG construction, which involves collecting and organizing data from heterogeneous sources such as databases, documents, or external data providers. Once the KG is constructed, it can be refined through machine learning and manual curation. This process relies on reasoning algorithms that can not only identify and correct inconsistencies in the data, but also add new information to the graph. Finally, the KG can be used for various NLP tasks to improve the efficiency and effectiveness of an enterprise's data management and decision-making processes.

Another observation from the presented use cases is the capability of KGs to capture a great variety of domain knowledge in an expressive and flexible fashion. The graph representation of nodes and edges is especially useful for areas with highly interconnected data items, which is more difficult to manage in traditional, relational databases. In our survey, we identified 20 different application domains. A subset of the most prominent ones can be seen in the left bar chart in Figure 15.2. Apart from the dominating health domain, there are many enterprise use cases emerging from other domains such as business, engineering, law, or energy. In particular, our investigation of use cases included topics about patents, fossil fuel exploration and production, finance, software engineering, human resources, product design, and many more. This diversity shows that KGs and NLP can be valuable for a very wide range of domains and applications.

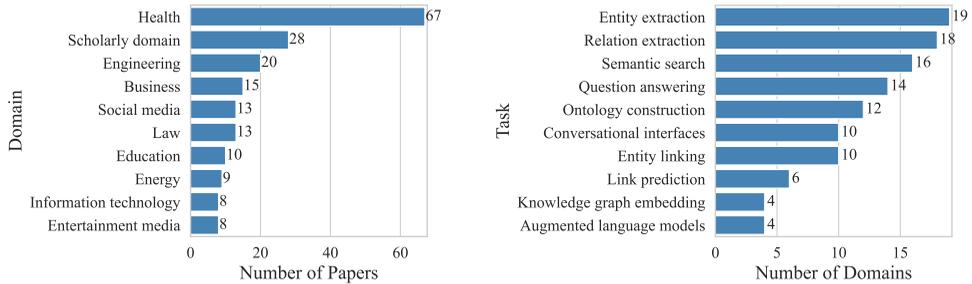

Figure 15.2: Number of papers by most popular application domains (left) and overview of most popular tasks by number of application domains (right) (Schneider et al. 2022a).

In view of the large number of tasks studied in the literature (see Figure 15.1), it is important to note that some are more mature in terms of practical applicability than others. According to our definition, maturity reflects the degree to which a given task has been examined with different research methods, resulting in a diverse set of available research contributions. Further, a given task can be considered more mature if it has been studied across a larger number of application domains, as depicted in the right bar chart in Figure 15.2. More mature tasks have a higher reliability of creating benefits in the discussed enterprise use cases. In line with the findings of our survey, tasks focused on KG construction as well as KG-based search systems are already mature (Schneider et al. 2022a). Accordingly, most papers from our analysis represent use cases for building up KGs from unstructured text or retrieving information from KGs, often in form of semantic search engines. Hence, the technology readiness of NLP-supported storage and access of data in KGs is reasonably high. While QA is an established task, there is a significant increase in research interest in the more advanced conversational interfaces. They can leverage the contextualized data in graphs to handle complicated information-seeking dialogues. However, more research work on deploying conversational agents and evaluating their usefulness in industrial settings is required. Other emerging research streams around KG embedding or KG-augmented language models have a high potential for enterprise use cases, too, although there are extremely few studies that apply these tasks in practical scenarios. Consequently, use cases that rely on graph embeddings or augmented language models are still actively researched and thus less mature. This concerns, for example, automatic document generation, embedding-based trading recommendation, or systems that use reasoning to infer new knowledge.

## 15.5 Conclusion

We analyzed enterprise use cases regarding the combination of KGs and NLP. By modeling semantic relationships among key entities and leveraging unstructured text sources, KG-based NLP can provide valuable insights and support decision-making in various application scenarios. Building upon the findings of our previous survey, we introduced the three core areas of KG construction, reasoning, and KG-based NLP tasks. For each corea area, we presented selected studies that apply these tasks in an enterprise context, including domains such as engineering, law, publishing, or finance. In addition, we assessed the maturity of use cases in terms of practical applicability.

Overall, we expect that the rapidly growing body of literature on combining KG and NLP will continue to bring forth innovative technologies and scholars will continue to explore even more application domains. Other enterprises will start to adopt KGs as a strategic tool to drive digital transformation, following the example of big technology companies like Google, which have harnessed the advantages of KGs for their internal processes and products for almost a decade now. Nonetheless, despite the strong business use cases, we observed a lack of applied research in practice. Compared to the whole body of literature about KGs in NLP, papers that focus on concrete enterprise use cases are still scarce. This might seem counterintuitive because the research interest in academia about KGs has been driven by the industry to a large extent. Hence, we call for more evaluation research in real-world scenarios that shed light on the added value of implemented solutions. This is crucial to make significant strides in advancing the field and transferring findings from academia to practical applications.